\documentclass[letterpaper, 10 pt, journal, twoside]{IEEEtran}

\usepackage{cite}
\usepackage{comment}
\usepackage{todonotes}
\usepackage{amsmath,amssymb,amsfonts}
\usepackage{booktabs}
\usepackage{graphicx}
\usepackage{textcomp}
\usepackage{xcolor}
\def\BibTeX{{\rm B\kern-.05em{\sc i\kern-.025em b}\kern-.08em
    T\kern-.1667em\lower.7ex\hbox{E}\kern-.125emX}}
    
\usepackage{paralist}
\usepackage[shortlabels]{enumitem}

\usepackage[ruled]{algorithm}
\usepackage{etoolbox}
\usepackage[noend]{algpseudocode}
\usepackage{MnSymbol}
\usepackage{siunitx}
\usepackage{hyperref}
\makeatletter
\def\BState{\State\hskip-\ALG@thistlm}
\makeatother

\usepackage{microtype}
\usepackage{makecell, tabularx}

\setcellgapes{2pt}
\newcolumntype{L}{>{\raggedright\arraybackslash}X}
\usepackage{caption}
\usepackage{changepage}

\begin{document}

\title{Maximising Coefficiency of Human-Robot Handovers through Reinforcement Learning}

\author{Marta Lagomarsino,
Marta Lorenzini,
Merryn Dale Constable,
Elena De Momi, \\ 
Cristina Becchio, and
Arash Ajoudani%
\vspace{-0.5cm}
\thanks{Manuscript received: January, 19, 2023; Revised March, 17, 2023; Accepted May, 8, 2023. This paper was recommended for publication by Associate Editor E. Yoshida and Editor A. Bera upon evaluation of the reviewers' comments. This work was supported by the ERC-StG Ergo-Lean (Grant No. 850932) and The Royal Society (Grant No. IES$\backslash$R3$\backslash$203086).
The authors thank Dr. M. Memeo, Dr. J.W. Ashmore Strachan, and M.Leonori for their help in the experiments.} 
\thanks{Corresponding author's email: {\tt\footnotesize marta.lagomarsino@iit.it}}
\thanks{M.Lagomarsino is with the Human-Robot Interfaces and Interaction Laboratory, Istituto Italiano di Tecnologia, Genoa, Italy, and also with the Department of Electronics, Information and Bioengineering, Politecnico di Milano, Milan, Italy.}
\thanks{M.Lorenzini and A.Ajoudani are with the Human-Robot Interfaces and Interaction Laboratory, Istituto Italiano di Tecnologia, Genoa, Italy.}%
\thanks{E.De Momi are with the Department of Electronics, Information and Bioengineering, Politecnico di Milano, Milan, Italy.}%
\thanks{M.D.Constable is with the Department of Psychology, Northumbria University, Newcastle, United Kingdom.}%
\thanks{C.Becchio is with the Department of Neurology, University Medical CenterHamburg-Eppendorf, Hamburg, Germany and Cognition, Motion and Neuroscience Laboratory, Istituto Italiano di Tecnologia, Genoa, Italy.}%
}

\markboth{IEEE Robotics and Automation Letters. Preprint Version. Accepted May, 2023}
{Lagomarsino \MakeLowercase{\textit{et al.}}: Maximising Coefficiency of Human-Robot Handovers} 



\maketitle

\IEEEpeerreviewmaketitle

\begin{abstract}
Handing objects to humans is an essential capability for collaborative robots. 
Previous research works on human-robot handovers focus on facilitating the performance of the human partner and possibly minimising the physical effort needed to grasp the object. However, altruistic robot behaviours may result in protracted and awkward robot motions, contributing to unpleasant sensations by the human partner and affecting perceived safety and social acceptance. 
This paper investigates whether transferring the cognitive science principle that ``humans act \textit{coefficiently} as a group'' (i.e. simultaneously maximising the benefits of all agents involved) to human-robot cooperative tasks promotes a more seamless and natural interaction. 
Human-robot \textit{coefficiency} is first modelled by identifying implicit indicators of human comfort and discomfort as well as calculating the robot energy consumption in performing the desired trajectory. We then present a reinforcement learning approach that uses the human-robot \textit{coefficiency} score as reward to adapt and learn online the combination of robot interaction parameters that maximises such \textit{coefficiency}. 
Results proved that by acting \textit{coefficiently} the robot could meet the individual preferences of most subjects involved in the experiments, improve the human perceived comfort, and foster trust in the robotic partner.
\end{abstract}


\section{INTRODUCTION}
\label{sec:introduction}

\IEEEPARstart{U}{nstructured} environments such as factories without work cells, households, and hospitals, where robots have the potential to assist humans, often involve robot-to-human handovers.
Effective handovers are not limited to the accurate and precise transfer of objects from the robot to the human but require physical and cognitive coordination. 
Previous studies have proposed methods whereby the robot facilitates human action by reducing physical effort. According to various human ergonomic metrics (e.g. distance to a neutral position, overloading joint torque, posture-based observational methods), the robot adjusted the position \cite{Bestick2018Learning, Kim2019adaptable} and orientation \cite{Busch2017postural, Calmak2021affordance} of the object, 
and learned its optimal location in space \cite{Mainprice2012sharing}, 
whole-body configuration \cite{Cakmak2011human}, 
and accomplished trajectory \cite{Sisbot2012ahuman}. 

\begin{figure}[!t]
    \centering
    \vspace{-0.4cm}  
    \includegraphics[width=0.78\linewidth]{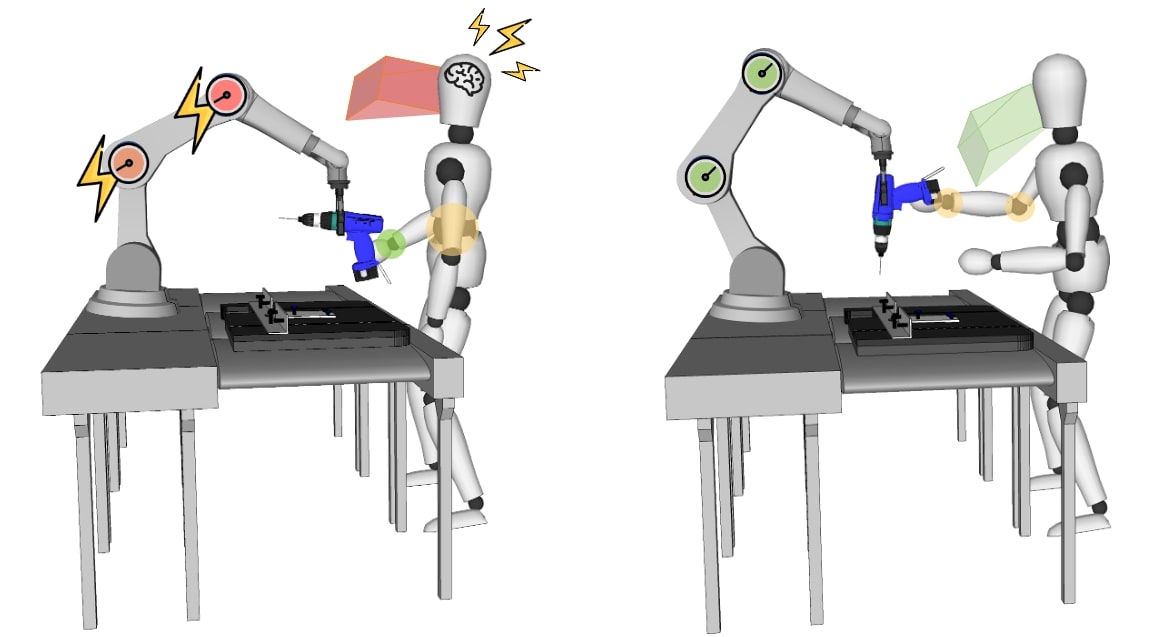}
    \vspace{-0.2cm}  
    \caption{\small 
    Illustration of the rationale behind human-robot \textit{coefficient} actions. Maximisation of human physical comfort may require awkward robot motion affecting perceived safety. Contrarily, considering efforts of both agents, a more seamless interaction can be achieved.}
    \vspace{-0.7cm}    \label{fig:conceptual_figure}
\end{figure}

Nevertheless, to achieve seamless human-robot interactions, it is preferable that robots also display understanding toward the socio-cognitive aspects of the interaction and aim at matching human preferences and skills \cite{Ortenzi2021object, lorenzini2023ergonomic}.
The socio-cognitive patterns of the interaction are as important as the physical ones for robots to be considered partners and not only tools. For example, imagine a robot handing over a drill to a human co-worker in a factory. The way the robot grasps and configures the object in the operating space affects the user's comfort, how convenient the tool's physical transfer is, and how efficiently the user can accomplish the subsequent action. An extreme maximisation of the human physical convenience could result in protracted and unnatural robot motions and negatively affect perceived safety and social acceptance \cite{Norman2005human} (see Fig.\ref{fig:conceptual_figure}). 
Although works on planning more legible and predictable robot motions exist in the literature \cite{Dragan2013legibility, Stulp2015facilitating}, gaps were identified in relation to holistic approaches, which examined all the aspects (socio-cognitive and physical) and phases (of the handover process) of human-robot interaction through a unified lens.

Cognitive science studies investigating human-human joint actions have highlighted that people are sensitive to the aggregate physical and cognitive effort of the dyad and tend to act \textit{coefficiently} as a group \cite{Torok2019rationality}. 
In other words, when cooperating with others to reach a shared goal, people consider the dyadic interaction as a whole and select actions that maximise the overall efficiency of the joint action rather than any individual components \cite{Strachan2020efficiency, Torok2021computing}. 

The present study aims to enable robots to learn to make \textit{coefficient} decisions, as humans generally do in human-human interactions. Indeed, robot behaviours that are more natural or human-like are preferred inasmuch they are more readable and foster trust in the human partner \cite{Cakmak2011human}.  
We propose an innovative approach whereby the robot online assesses the comfort of the specific human it interacts with, both at the socio-cognitive and physical levels, and simultaneously takes into account its internal expenses (e.g. motion feasibility and energy consumption). Thereby, the robot uses this information to choose a handover configuration that results in \textit{coefficient} actions for the two agents (i.e. concurrently efficient for the robot passer and the human receiver). 

The human-robot \textit{coefficiency} is modelled by online capturing implicit comfort and discomfort body signals from the human partner as well as the robot energy consumption. Specifically, for the former, we estimate human cognitive and physical ergonomics during the interaction by analysing the reaction time, the attention distribution, and the upper-body kinematics.  
This allows us to estimate the aggregate expenses of the dyad during the interaction, define a human-robot \textit{coefficiency} score, and learn through a reinforcement learning (RL) approach the actions that maximise such \textit{coefficiency}. 
At each robot-to-human handover iteration, the robot explores different values of the considered interaction parameters, i.e.
\begin{inparaenum}[(i)]
    \item the object orientation, 
    \item the interaction distance, and
    \item the velocity in approaching the human partner. 
\end{inparaenum} 
It reads the reward obtained, which is based on the aforementioned human-robot \textit{coefficiency} score. Then, it decides whether to exploit the gathered information to maximise the short-term reward (by selecting the subsequent interaction parameters accordingly) or keep exploring the environment.  

Nevertheless, planning robot motions that match human preferences is tricky. 
Learning biologically inspired robot trajectories is often not enough since human preferences are profoundly subjective and often vary when familiarising the task.  
For example, it has been shown that the comfortable distance perceived by each specific-subject changes with the feeling of menace in the robot actions and the smoothness of the accomplished trajectory \cite{Lagomarsino2022robot}. 
Hence, we design a system that does not learn each interaction parameter separately but takes into account the considered interaction parameters simultaneously to ultimately find the combination that best fits users' personal preferences.

To the best of our knowledge, this is the first time that the human tendency to act \textit{coefficiently}, which differs from altruistic behaviour commonly adopted in the literature \cite{Bestick2018Learning, Busch2017postural}, is designed and developed in human-robot handovers. 
The proposed handover learning and adaptation system is tested on twelve subjects in a daily collaborative activity, where the robot hands over a mug to the human counterpart to prepare some coffee (see multimedia attachment\footnote{The video can also be found at \href{https://youtu.be/VYwnkW5AIJU}{youtu.be/VYwnkW5AIJU}.}).

\section{
Human-Robot Coefficiency Model
}
\label{sec:coefficiency}
In this section, we define a set of variables to model \textit{coefficiency} in human-robot joint actions, such as handovers, and describe how to evaluate it online without affecting the natural flow of the interaction. 

Regarding human contribution, we investigate variables related to human cognitive and physical ergonomics during task execution. 
The only requirement here is the online collection of data about the object position and human motion through a suitable sensor system (e.g. visual tracking or IMU-based motion capture systems).  
The musculoskeletal system of the human body can be modelled by $N$ articulations (joints) and $N+1$ body segments (rigid links). 
A frame $\Sigma_i$ is associated with each joint $i \in \{1, \dots, N\}$ and its configuration over time with respect to a world frame $\Sigma_\text{W}$ is known thanks to an initial calibration. Each joint can feature $D\leq3$ degrees of freedom (DoFs), and the angle with respect to its parent link is denoted by $q^\text{H}_{i} \in \mathbb{R}^{D}$. The kinematic chain imposes a set of constraints on the link's motion patterns, which leads to the definition of the joints Range of Motion (RoM):
\begin{equation}
\label{eq:rom}
    q^\text{H}_{i,k,\text{min}} < q^\text{H}_{i,k} < q^\text{H}_{i,k,\text{max}}, \,\, i \in \{1,\dots, N\}, \,k \in \{1, \dots, D\}
\end{equation}

For the robot, we measure the energy consumed to complete the planned trajectory. We consider a robotic manipulator constituted by a serial collection of rigid links, which are connected by $M$ revolute joints exhibiting one DoF each (represented by the angle $q^\text{R}_{j} \in \mathbb{R}, j \in \{1,\dots, M\}$). 

\vspace{-0.1cm}
\subsection{Human Cognitive Ergonomic Cost}
\label{sec:cognitive}

Regarding the social-cognitive level of the interaction, we analyse the reaction time $\tau$ and the attention that the human receiver gives to the object that he/she has to handle.
Indeed, studies on the control of human body motion in social contexts highlight that human actions requiring a more significant amount of planning result in motion initiation latencies \cite{Khan2006inferring}. To detect changes in the kinematics of human movements during human-robot interaction, we consider the time elapsed between the time instant in which the robot starts its motion $t^\text{R}_0$ and the human motion initiation time $t^\text{H}_0$, and we normalise the value to the total execution time $\Delta t$ of the robot's trajectory
\vspace{-0.1cm}
\begin{equation}
    \tau = \dfrac{t^\text{H}_0 - t^\text{R}_0}{\Delta t}.
\vspace{-0.1cm}
\end{equation}
Standardising reaction times against the duration of the observed movement is a common practice when considering joint actions with movements of different duration \cite{Podda2017theheaviness}.

Moreover, behavioural and neuroscientific studies have provided evidence that discomfort and cognitive load usurp executive resources, which otherwise could be used for attentional control, thus increase distraction \cite{Lagomarsino2022anonline}. 
To estimate the level of attention toward the task, we consider the head frame, translate it in correspondence to the centre of the head link and tilt it ten degrees to approximate the gaze direction \cite{Weidenbacher2007acomprehensive}
(denoted as $\Sigma_\text{gaze}$ from now on, see Fig.\ref{fig:framework}). Consequently, the Cartesian vector expressing the relative position between $\Sigma_\text{gaze}$ and $\Sigma_\text{object}$, namely the frame associated with the object that should be handled, is mapped into spherical coordinates (azimuth angle $\theta$, elevation angle $\varphi$ and radial distance). 
A fuzzy logic membership function exploiting the Raised-Cosine Filter \cite{Lagomarsino2022anonline} is then applied to normalise the measured attention angles at each time instant $t$ ($\theta(t)$ and $\varphi(t)$ angles, indicated in Eq.(\ref{eq:attention}) as $\alpha(t)$) in the range $[\alpha_{\text{min}}(t),\alpha_{\text{max}}(t)]$. 
\begin{equation}
\label{eq:attention}
\!\! f(\alpha{\scriptstyle (t)})\! =\!
\begin{cases}
    1, 
        & {\displaystyle \!\!\!\! \mbox{if } \! \left\lvert\alpha{\scriptstyle (t)}\right\rvert \leq\alpha_{\text{min}}{\scriptstyle (t)}} \\
        
    \dfrac{1}{2} \Bigl[1 \!
        - \cos{\Bigl(\dfrac{\left\lvert\alpha{\scriptstyle (t)}\right\rvert - \alpha_{\text{min}}{\scriptstyle (t)}}
        {\alpha_{\text{max}}{\scriptstyle (t)} - \alpha_{\text{min}}{\scriptstyle (t)}}
        \pi\Bigr)} \Bigr], 
        & {\displaystyle \!\!\!\!\! \underset{\displaystyle \& \left\lvert\alpha{\scriptstyle (t)}\right\rvert \leq \alpha_{\text{max}}{\scriptstyle (t)}}{\mbox{if } \!   \left\lvert\alpha{\scriptstyle (t)}\right\rvert\! 
        >\alpha_{\text{min}}{\scriptstyle (t)}
        }} \\
    0, 
        & {\displaystyle \!\!\!\!
        \mbox{otherwise.}}
\end{cases}
\end{equation}
Note that the threshold values on $\alpha{(t)}$ (i.e. control points $\alpha_{\text{min}}{(t)},\alpha_{\text{max}}{(t)}>0$) depend on the current distance of the moving object from the human operator.
Specifically, the lower limit $\alpha_{\text{min}}(t)$ and upper limit $\alpha_{\text{max}}(t)$ are computed as 
\vspace{-0.2cm}
\begin{equation}
\label{eq:attention_limits}
    \alpha_{\text{min}}{(t)}=\text{tan}^{-1}\!\left(\frac{(1-\gamma) \,r}{d{\scriptstyle (t)}}\right), 
    \quad
    \alpha_{\text{max}}{(t)}=\text{tan}^{-1}\!\left(\frac{(1+\gamma) \,r}{d{\scriptstyle (t)}}\right),
\vspace{-0.1cm}
\end{equation}
where $r$ is the radius of the area in which the object is assumed to be contained (i.e. the fixed distance between the origin of the frame $\Sigma_\text{object}$ associated with the object and its boundary), while $d(t)$ is the distance between the origins of $\Sigma_\text{gaze}$ and $\Sigma_\text{object}$. 
$\gamma$ is set to $0.4$ to determine a smooth behaviour (from $1$ to $0$) of the function $f(\alpha{\scriptstyle (t)})$ in the angle range corresponding to the distance range $[ (1-\gamma)\, r, \; (1+\gamma)\, r]$ with respect to the origin of $\Sigma_\text{object}$. 

The attention level $\Lambda(t)$ toward the task is therefore defined as the product between the normalised azimuth and elevation indicators and values of $\Lambda(t)$ closer to $1$ indicate a total focus on the region of interest
\vspace{-0.2cm}
\begin{equation}
    \Lambda(\theta{\scriptstyle (t)}, \varphi{\scriptstyle (t)})\, =  f(\theta{\scriptstyle (t)}) \, f(\varphi{\scriptstyle (t)}).
\end{equation} 

\subsection{Human Physical Ergonomic Cost}
\label{sec:physical}
The physical comfort is monitored by considering the joints RoM values (see Eq.(\ref{eq:rom})). We take inspiration from ``pen-and-paper'' ergonomics indexes, such as Rapid Upper Limb Assessment (RULA)\cite{mcatamney1993rula} and Rapid Entire Body Assessment (REBA)\cite{hignett2000rapid}, 
which claim that a human is exposed to physical effort if one of his/her joints is close to the RoM extrema. Similarly to \cite{Gholami2022quantitative}, we parametrise the ergonomic cost for each $k$-th DoF of $i$-th joint at instant $t$ as
\begin{equation}
    \zeta_i^k(t) = 
    \dfrac{2\min\big\{\left\lvert q^\text{H}_{i,k} - q^\text{H}_{i,k,\text{min}} \right\rvert,
    \left\lvert q^\text{H}_{i,k} - q^\text{H}_{i,k,\text{max}} \right\rvert \big\}}
    {\left\lvert q^\text{H}_{i,k,\text{max}} - q^\text{H}_{i,k,\text{min}} \right\rvert}.
\end{equation}
Note that $\zeta_i^k(t)$ spans in the interval $[0,1]$, and the closer the value is to 1.0, the more comfortable posture the human is experiencing.
Then, we identify the most stressed DoF for each joint and we average the stress effects over all $N$ joints 
\vspace{-0.2cm}
\begin{equation}
    \bar{\zeta}(t) = \frac{1}{N}\sum_{i=1}^{N} \, \Big[ \min_{k=1,\dots D}{\zeta_i^k(t)} \Big].
\end{equation}

\subsection{Robot Consumption Cost}
\label{sec:robotcost}
The robot efficiency is parametrised in this work by the robot power consumption \cite{Mohammed2014minimizing}. Indeed, the latter is sensitive to robot behaviours that could affect human comfort (and thus would like to avoid), e.g. unnatural and too fast motions (resulting in higher torques and velocities, respectively). 
At a specific instant $t$, the power required to move the $j$-th robot joint is 
\vspace{-0.2cm}
\begin{equation}
    P_j(t) = \left\lvert \tau_j(t) \,\, \dot{q}^\text{R}_j(t) \right\rvert,
\end{equation}
\vspace{-0.5cm}

\noindent where $\tau_i(t)$ is the torque applied at $j$-th joint and $\dot{q}^\text{R}_j(t)$ is the $j$-th joint velocity. 
Summing up the contributions of all $M$ robot joints, we have
\vspace{-0.3cm}
\begin{equation}
    P(t) = \sum_{j=1}^{M} P_j(t).
\end{equation}

\subsection{Coefficiency of Human-Robot Joint Actions}

A human-robot \textit{coefficiency} score is associated with each conjoint action $a$ executed in $t \in [t_0, \dots, t_\text{f}]$ representing how efficient the latter is in terms of aggregate costs of the involved agents. 
More specifically, the score is modelled by integrating the quantities described in the above sections over the entire interaction duration (e.g. pre-handover phase, physical exchange and subsequent action) as follows
\begin{align}
\label{eq:coefficiency}
    C^\text{HR}_\text{coefficiency}(a) = \frac{1}{3} & \Big[ C_\text{cognitive erg}^\text{H} + C_\text{physical erg}^\text{H} + C_\text{energy cons}^\text{R} \Big]
\vspace{-0.2cm}
\end{align}
where $C_\text{cognitive erg}^\text{H}$ and $C_\text{physical erg}^\text{H}$ parametrise the human efficiency while $C_\text{energy cons}^\text{R}$ refers to the robot efficiency.  
In particular, the human cognitive ergonomic cost is defined as
\begin{equation}
\label{eq:cognitive_cost}
    C_\text{cognitive erg}^\text{H}(a) = \frac{1}{2} \Big[(1-\tau) + \underset{{t=t_0,\dots {t_\text{f}}}}{\mathbb{E}}[\Lambda(t)]\Big]. 
\end{equation}
where $\underset{{t=t_0,\dots {t_\text{f}}}}{\mathbb{E}}$ indicates the mean value of human attention level during the handover entire execution. 
This formulation is based on a study of human movements behaviour in human-robot interaction \cite{Kanda2003body}. The latter analysed human body movements in several aspects (such as eye contact, distance, synchronisation to the robot, and touches) and computed the correlations with subjective evaluations of robot behaviour. 
Results indicated that higher-ranked human-robot interactions were characterised by periods of intensive attention and motion synchronisation to the robot. In particular, multiple linear regression analysis revealed that these two aspects were the most relevant and equally significant ones (standardised partial regression coefficients were $0.476$ and $0.535$, respectively). 
Hence, taking inspiration from that study, we formulate our cognitive cost to be positively correlated to the average attention an individual gives to the task and negatively correlated with the reaction time (indeed the higher is $\tau$, the less synchronised the human is to the robot motion).

On the other hand, the physical ergonomic cost 
\begin{equation}
\label{eq:physical_cost}
    C_\text{physical erg}^\text{H} = \min_{t=t_0,\dots {t_\text{f}}} \bar{\zeta}(t)
\end{equation}
identifies the worst posture assumed during the interaction. We did not consider accumulative values since the interaction was quick, nor the averaging factor to avoid underestimating bad postures assumed for a short period of time. 

Finally, the robot efficiency is computed as the normalised energy consumption to execute the desired trajectory, e.g. 
\begin{align}
\label{eq:robot_cost}
    C_\text{energy cons}^\text{R}(a) = 1 - \dfrac{1}{E_\text{max}}\int_{t_0}^{t_\text{f}} P(t) dt .
\end{align}
The reader should note that the costs defined in Eq.(\ref{eq:coefficiency}), (\ref{eq:cognitive_cost}), (\ref{eq:physical_cost}), and (\ref{eq:robot_cost}) are normalised in $[0,1]$, and values of these indexes closer to $1$ denote high comfort for the agents.

\section{Handover Adaptation System}

\begin{figure}[!t]
    \centering
    \includegraphics[width=\linewidth]{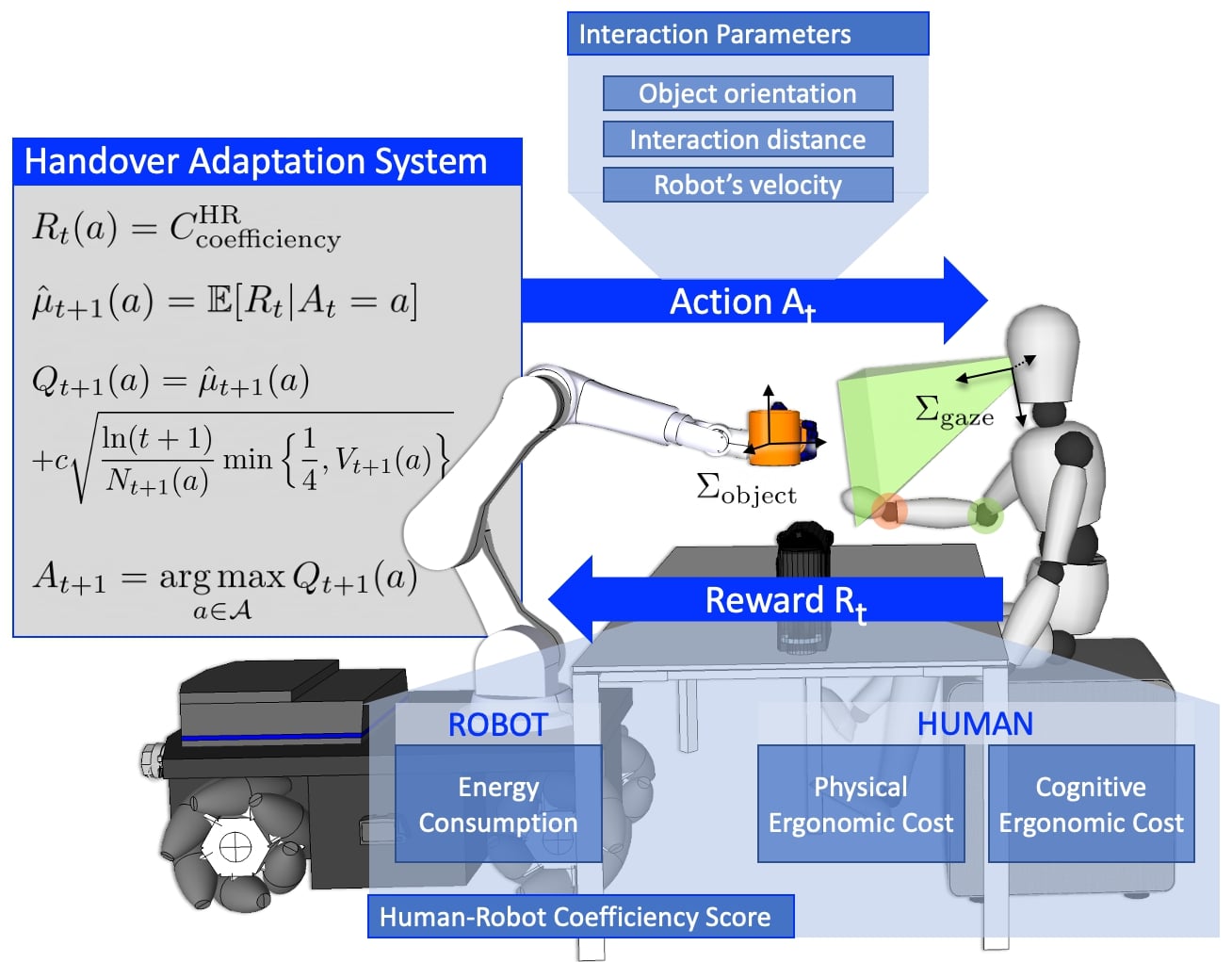}
    \caption{\small Overall structure of the proposed framework to transfer human paradigm of acting \textit{coefficiently} in human-robot handovers.}
    \label{fig:framework}
    \vspace{-0.5cm}
\end{figure}

Our robot behaviour adaptation system observes subconscious responses from the user revealing his/her comfort as well as the energy efficiency of the robot itself, and online adjusts the interaction parameters to maximise these monitored signals. The latter are combined in the human-robot \textit{coefficiency} score (described in the previous section) and then used as the reward that the system should maximise.
Exploiting a RL approach based on a Multi-Armed Bandit (MAB) algorithm, the robot explores and learns on the fly without separating the data collection and learning phases (system overview in Fig.\ref{fig:framework}).

\subsection{Adapted Parameters}
In this work, the adaptation of the handover strategy operates on three interaction parameters, i.e.
\begin{inparaenum}[(i)]
    \item the object orientation on the horizontal plane,
    \item the interaction distance, and
    \item the robot velocity profile.
\end{inparaenum} 
The selection of these parameters was motivated by their straightforward implementation, the desire to keep the dimensionality of the search space limited, and, above all, their irrefutable impact on the interaction. 
Indeed, how the robot positions and orients the transferring object determines the ergonomics of the physical exchange and how convenient is the completion of the subsequent action.  
However, human comfort deeply depends on person-specific physical characteristics such as the length of body segments and joints RoM. Moreover, the confidence level with technology influences human perceived comfort during robot motion and affects the amount of planning required by the human \cite{Lagomarsino2022pick}. 
Some subjects may prefer to incur a slightly extra physical effort if it would make the robot keep a distance perceived as safer\cite{Kirschner2022expectable}. 
These feelings also vary with the velocity profiles adopted by the robot to approach the human \cite{Kulic2007physiological}. 
Hence, to summarise, optimising the interaction to the user preferences requires considering all the combinations of these parameters simultaneously. 

\subsection{Observed Reward}
Despite the numerous studies in the literature, fully human-centred approaches are not guaranteed to be the best choice to achieve seamless human-robot interactions \cite{Norman2005human}. One could claim that the closer the object is and with the grasp affordance region more oriented to the user, the more comfortable the interaction is. However, an
extreme maximisation of the user's physical convenience could result in protracted and unnatural robot motions, contributing to unpleasant sensations by the human partner and affecting perceived safety and social acceptance. 
For this reason, in our RL scenario, the reward observed $R_t$ at each iteration $t$ keeps into account variables related to human ergonomics but also the convenience for the robot to execute the action. Namely, we considered 
\begin{equation}
    R_t(a) = C^\text{HR}_\text{coefficiency}(a). 
\end{equation}
The sensitivity to excessive robot expenses is motivated by the desire to enable natural and pleasant human-robot interactions. 

\subsection{Multi-Armed Bandit Problem}
\label{sec:mab}

An agent learning how to optimally interact with a new human partner quickly faces the exploration-exploitation dilemma. Namely, it must decide whether to continue exploring new actions or to perform the one that has earned it the highest rewards so far. 
Dealing with this trade-off efficiently is crucial in human-in-the-loop systems like ours, where testing time is limited.
To tackle this challenge, the RL community introduced the principle of \textit{optimism in the face of uncertainty}.  
This heuristic states that, despite the lack of knowledge about the environment, the agent makes an optimistic guess about how good the expected reward of each action is and selects the action with the highest guess. If the model is correct, the agent has no regrets, otherwise it updates its internal knowledge, diminishing the optimistic guess associated with that action and thus inducing the exploration of other actions.
As the agent resolves its uncertainty, the effects of optimism diminish, and the agent's policy approaches optimality.

Our work focuses on a finite-horizon MAB problem, i.e. a specific form of RL enabling the exploration-exploitation of the environment without changing the state. Other RL techniques would require defining a set of possible states and transition probabilities in a Markov decision process that can not be done for our application.  
More specifically, we consider a finite set of possible values for each parameter and define a $K$-armed bandit problem, where each arm corresponds to a robot action with a different combination of interaction parameters. 
At each iteration $t$, among the actions $\mathcal{A} \in \mathbb{R}^K$, the robot (as the agent) chooses an action (i.e. arm $a \in \mathcal{A}$) to perform and receives a reward $R_t(a)$. Then, the robot updates its internal knowledge about the expected reward 
\begin{equation}
    \hat{\mu}_{t+1}(a) = \hat{\mu}_{t}(a) + \dfrac{R_t(a) - \hat{\mu}_{t}(a)}{N_t(a)}.
\end{equation}
Note that the expected reward $\hat{\mu}_{t+1}(a)$ is no more than the 
average reward associated to the action $a$ estimated iteratively on the basis of the observed reward $R_t(a)$ and the number $N_t(a)$ of times $a$ was taken prior to $t$. This formulation is memory efficient since it avoids storing all past rewards and re-calculating the average at each time step.

To improve the robot policy required to select the subsequent action, we use the Upper Confidence Bound (UCB) algorithm \cite{Auer2002finite} that
asymptotically achieves the logarithmic regret\footnote{The regret for a policy is defined as the difference between the reward obtained and the highest expected reward.}. 
For each action $a$, we compute UCB1-tuned value 
\begin{equation}
\label{eq:qvalue}
    Q_t(a) = \hat{\mu}_{t}(a) + c \sqrt{\dfrac{\text{ln}(t)}{N_t(a)} \min\Big\{\frac{1}{4}, V_t(a)\Big\}},
\end{equation}
where the second term denotes the confidence level of the estimate ($c>0$). $V_t(a)$ is the upper confidence bound on the variance of the action $a$, computed on the basis of the rewards obtained until $t$, 
\vspace{-0.2cm}
\begin{equation}
    V_t(a) =  \sum_{k=\{t|A_k=a\}}\dfrac{\hat{\mu}_{k}(a)^2}{N_t(a)} - \hat{\mu}_{t}(a)^2
    + \sqrt{\dfrac{2 \, \text{ln}(t)}{N_t(a)}},
\vspace{-0.1cm}
\end{equation} 
and the factor $1/4$ is the upper bound on the variance of a Bernoulli random variable. This means that the action $a$, which has been played  $N_t(a)$ times during the first $t$ plays, has a variance that is at most the sample variance plus $\sqrt{2 \, \text{ln}(t)/N_t(a)}$.
On each subsequent pull, the agent picks the action $A_{t}$ that maximises $Q_t(a)$, namely
\begin{equation}
    A_{t} = \underset{a\in \mathcal{A}}{\text{arg\,max}} \, Q_{t}(a).
\end{equation}

It can be noticed that UCB moves from focusing primarily on exploration (when the actions that are tried the least are preferred) to instead concentrating on exploitation (selecting the action with the highest estimated rewards).

\section{Experiments}
\label{sec:experiments}

\begin{figure}[!t]
    \centering
    \includegraphics[width=0.9\linewidth]{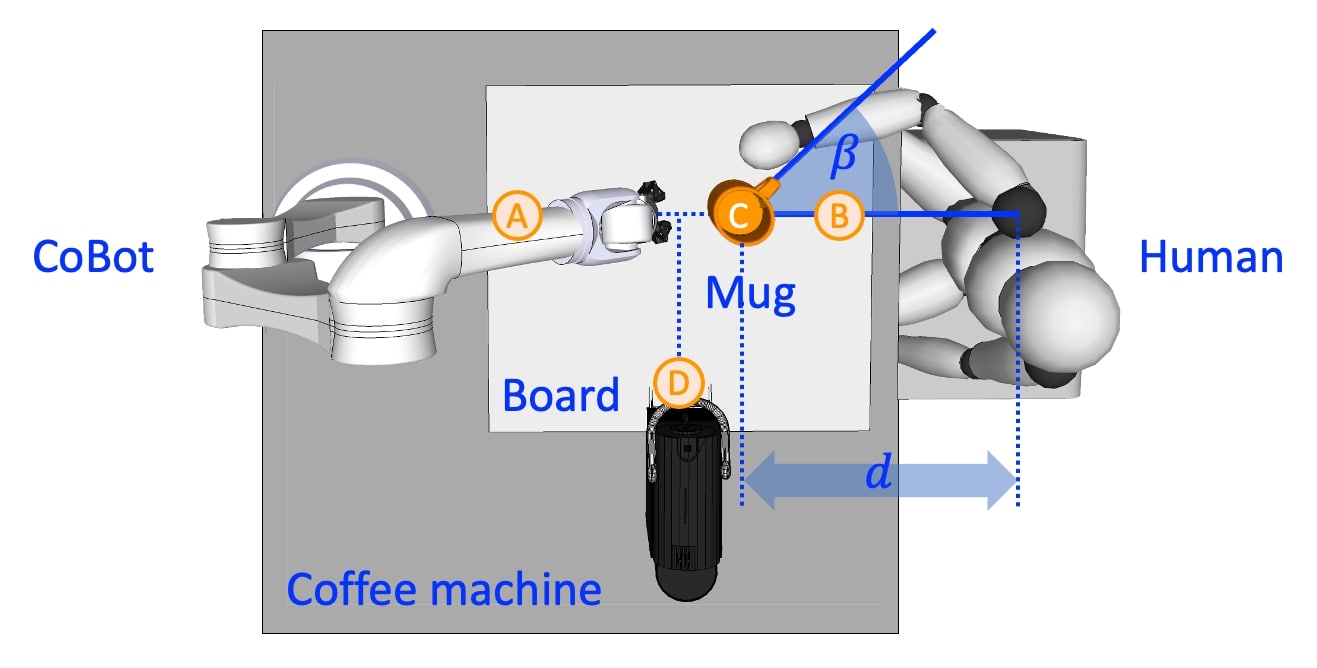}
    \vspace{-0.3cm}
    \caption{\small Overview of the experimental setup involving collaborative robot, human partner, and object for handover (i.e. mug).
    }
    \label{fig:setup}
    \vspace{-0.4cm}
\end{figure}

In this section, we describe an experimental analysis conducted to assess the capability of the proposed framework in learning to make \textit{coefficient} decisions and evaluate its potential for improving human-robot interaction\footnote{The experiments were carried out at HRII Lab, in accordance with the revised Helsinki Declaration, and the protocol was approved by the ethics committee Comitato Etico Regionale della Liguria (IIT\_ERC\_IMOVEU version 03.1 29/06/2022).}. 
More specifically, two research questions (RQs) were tested: 
\begin{enumerate}[RQ1., leftmargin=7.5mm]
    \item \textit{Are the interaction parameters learned by our framework resulting in efficient actions for the involved agents?} \label{RQ1}
    \item \textit{Does the proposed coefficiency-based decision-making strategy allow aligning the robot behaviour to the preferences of the human partner?} \label{RQ2}
\end{enumerate} 
Note that preferences are meant for interaction parameters, and perceived naturalness, appropriateness and trust. 

\subsection{Experimental Protocol}
Twelve healthy subjects, three men, eight women, and one non-binary ($26.1\pm3.3$ years), were recruited in the experiments. Participants signed written informed consent, declared to have never interacted with a manipulator before and were na$\ddot{\i}$ve to the experimental purpose.
We considered a scenario where a robot performs a day-to-day task next to a person (setup in Fig.\ref{fig:setup}). Specifically, the collaborative robot (Franka Emika Panda) picked a mug from position \textit{A} and handed it to the human (in \textit{C}) sitting at the table. The latter, starting from \textit{B}, grasped the object and placed it under the coffee machine (i.e. \textit{D}).
We asked participants to repetitively perform the action fifty times in the most natural way as they would not be observed. 
To measure the kinematics of human motion, we exploited a wearable MVN Biomech suit (Xsens Tech.BV) based on inertial measurement unit sensors. For this study, we restricted the analysis of physical ergonomics to the right wrist and elbow joints since they are the most involved in this task. Note that all participants were right-handed. Besides, the object position was estimated online by considering a fixed transformation from the robot grasping position during the handover phase and from the tracked human wrist during the post-handover action. 
A board with buttons in the locations mentioned above was also designed to assess human motion initiation time and thus the reaction time $\tau$ more precisely. Indeed, participants were trained to keep pressing the button until the action of grasping the mug. 

\subsection{Parameters Adaptation}

The robot ran an impedance controller and tracked trajectories computed by smoothly interpolating a sequence of desired configurations.  
Different robot behaviours were implemented by adapting online the performed trajectory. 
The trajectory starting point was fixed to the robot configuration to grasp the mug on the table in \textit{A}. The following poses the robot passes through and the associated timing law varied according to the parameters learned by the adaptation system. 
The participants experienced three different final object orientations ($\beta_1=\pi/6$, $\beta_2=\pi/2$ and $\beta_3=5\pi/6$, obtained by evenly sampling the range identified in \cite{Constable2016ownership} by the final angle a human passer places the handle to facilitate the grasp of a human receiver, see Fig.\ref{fig:setup}), two interaction distances ($d_1=0.30\si{\meter}$ and $d_2=0.45\si{\meter}$, i.e. the middle and the limit of the \textit{intimate distance} range proposed in \cite{Kirschner2022expectable}) and two total execution times of the robot trajectory ($\Delta t_1=5.0\si{\second}$ and $\Delta t_2=8.0\si{\second}$, in line with natural and fast human movements registered in \cite{becchio2007different}).
Thus, a twelve-armed bandit problem based on UCB1-tuned is defined as presented in Sec.\ref{sec:mab}. 
The confidence value was set to infinity ($c=+\infty$) for the unexplored arms, inducing an initial priming round to be performed, in which each action $a$ was sampled once to obtain the initial value of $\hat{\mu}_t(a)$. This avoided divide-by-zero errors in the exploration term $Q_t(a)$ when actions have not yet been tried and $N_t(a)$ is equal to zero. 
The policy then explored with $c=0.1$, reduced the uncertainty (decrease of second term in Eq.\ref{eq:qvalue}) and learned the optimal combination of parameters for the specific user the robot is collaborating with. 
For the test, $E_\text{max}$ was set to the maximum energy consumed among the proposed trajectories.

\subsection{Subjective Questionnaires}

After the experiment, we asked participants to select the parameters they felt more comfortable for the interaction (i.e. preferred object orientation, distance and robot velocity). Moreover, they ranked the naturalness and seamlessness of the handover technique using five-point Likert scale questions (see Table \ref{tab:questionnaires} where scoring $1$ indicates that the subject strongly disagrees and $5$ strongly agrees with the statement). 
The evaluation included a technique developed by NASA to assess the relative importance of factors in determining the final score, i.e. the coefficients of their combination. Pairs of costs involved in the \textit{coefficiency} score were presented, and subjects were asked to select which of the two should be taken more into consideration while planning robot motions. From the pattern of choices, we computed the weights that each subject would associate with costs in Eq.\ref{eq:coefficiency}.  
A copy of the custom questionnaire can be found as supplementary materials for this paper. 
Finally, they filled the trust scale defined in \cite{Charalambous2016}\footnotemark to assess the perceived appropriateness of the robot motion and pick-up speed, cooperation safety, and reliability. 

\begin{figure}[!t]
    \centering
    \includegraphics[width=\linewidth]{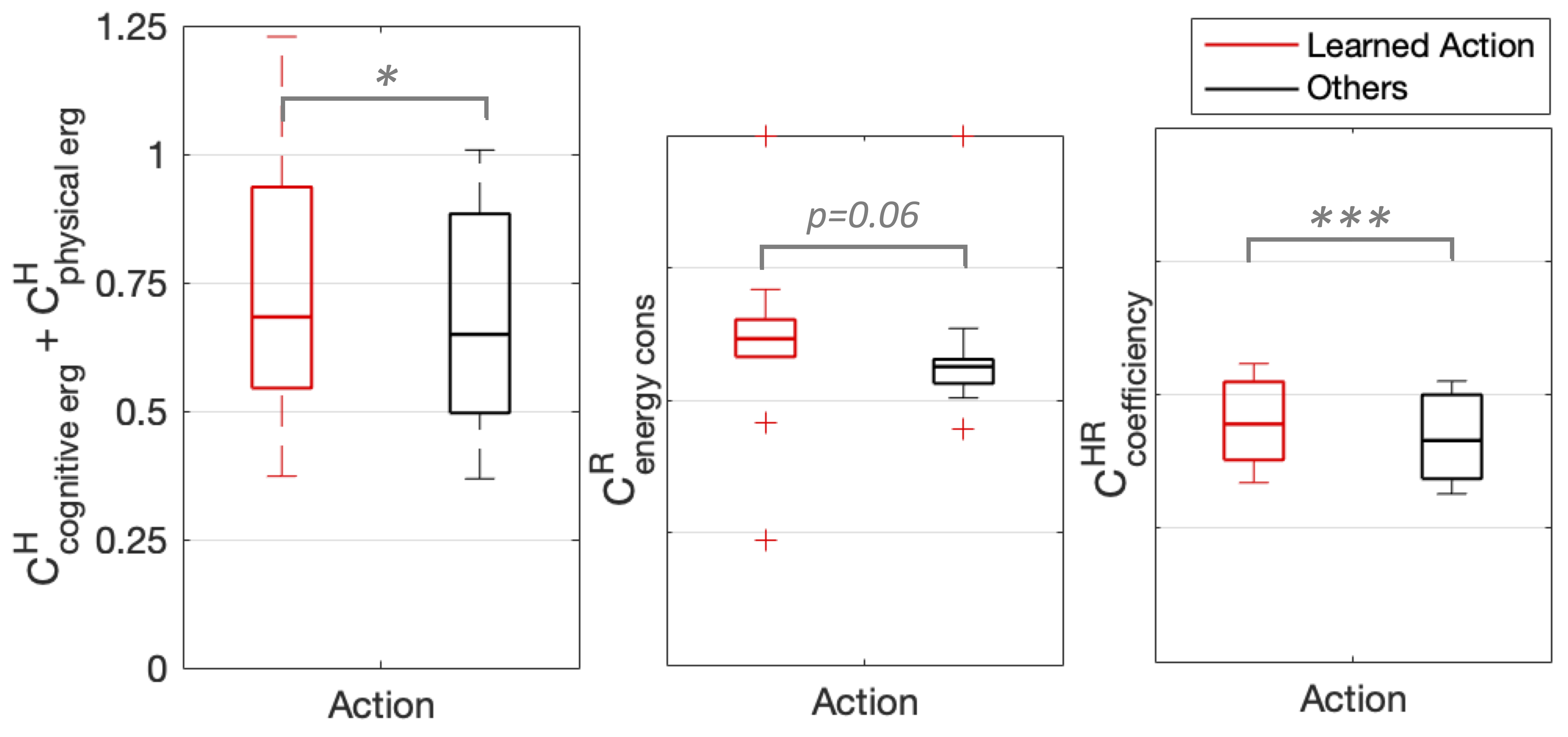}
    \vspace{-0.6cm}
    \caption{\small Comparison of human ergonomics, robot energy consumption and human-robot \textit{coefficiency} score running the action learned by the framework and all the other iterations. 
    } 
    \label{fig:statistic_analysis2}
    \vspace{-0.5cm}
\end{figure}

\begin{figure}[!t]
    \centering
    \includegraphics[width=\linewidth]{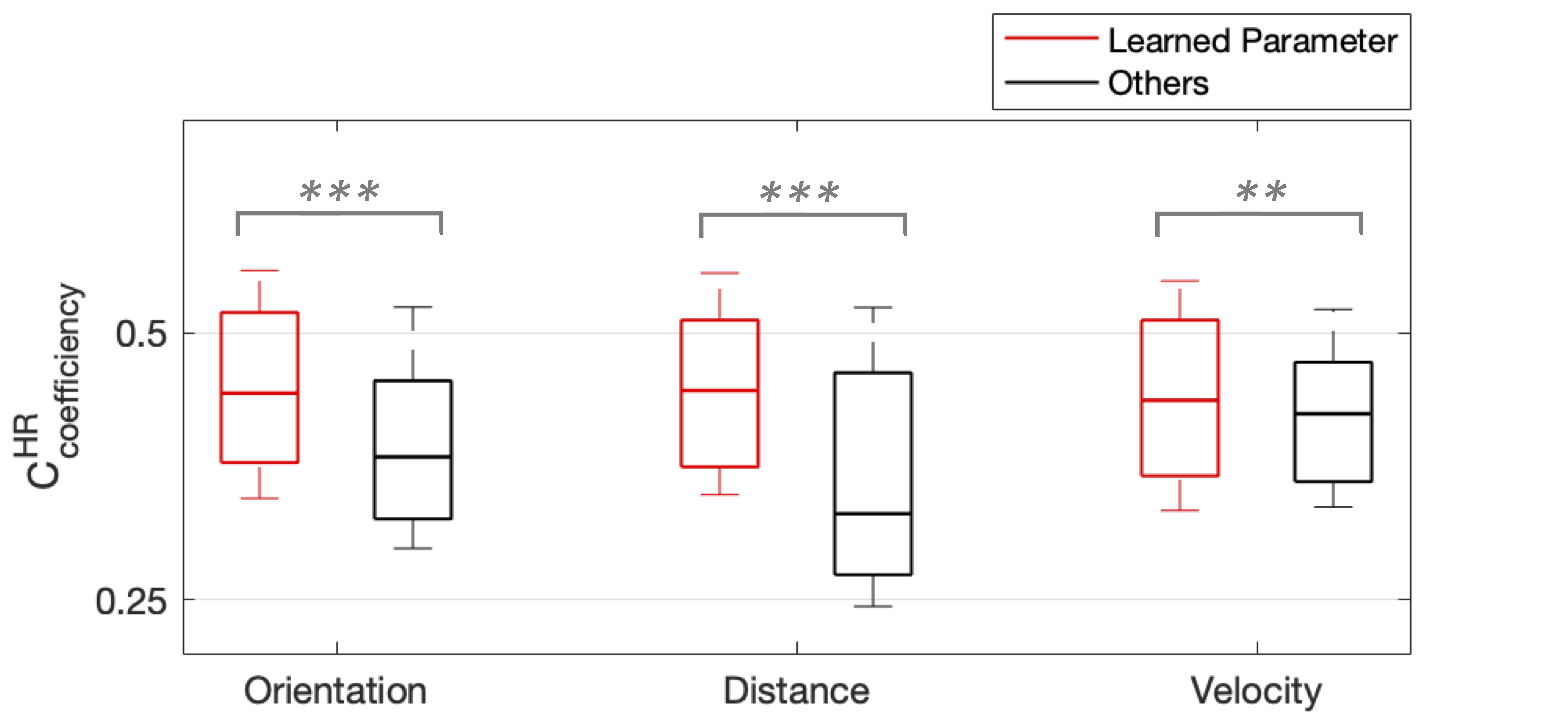}
    \vspace{-0.6cm}
    \caption{\small Comparison of human-robot \textit{coefficiency} score obtained exploiting a specific interaction parameter value learned by the framework and all the other iterations. 
    Wilcoxon test significance levels are indicated at *p\textless{}0.05, **p\textless{}0.01, ***p\textless{}0.001. 
    }
    \label{fig:statistic_analysis}
    \vspace{-0.6cm}
\end{figure}

\section{Experimental Results}
\label{sec:result}
\subsection{Learning and Adaptation Results}

A statistical analysis using the non-parametric Wilcoxon signed-rank test (WSRT) was conducted to compare the efficiency metrics computed when the robot exploited the learned parameters and in all other iterations (\ref{RQ1}). 
As can be seen in Fig.\ref{fig:statistic_analysis2}, overall, the efficiency of the actions performed by both the involved agents improved in value thanks to the learning.
\footnotetext{Note that the authors of \cite{Charalambous2016} provide the average outcome and standard deviation of each questionnaire item following the experimental trials on one hundred and fifty-five subjects at Cranfield and Loughborough Universities. 
}
A significant increase in the human ergonomic cost (as the sum of $C^\text{H}_\text{cognitive erg}$ and $C^\text{H}_\text{physical erg}$) was registered when executing the optimal action learned by the system for each specific user ($\text{p}^{A}=0.027$)\footnote{ $\text{p}^V$ denotes the p-value between iterations exploiting learned values and all the others and $V$ is a specific interaction parameter, i.e. orientation $\beta$, distance $d$ or velocity $\Delta t$, or a combination of them, i.e. an action $A$.}. 
Moreover, the human-robot \textit{coefficiency} cost experienced a growth of $10.6\%$ in the median ($\text{p}^{A}<0.001$).  
From Fig.\ref{fig:statistic_analysis}, we can also notice a significant effect of each interaction parameter on the reward of our RL algorithm. Indeed, the mug's orientation and distance learned by the presented policy predominately increased the \textit{coefficiency} of the human-robot dyad ($\text{p}^{\beta}$, $\text{p}^{d}<0.001$). 
The same can be stated for the robot velocity ($\text{p}^{\Delta t}=0.002$) although with lower significance.

\begin{figure}[!t]
    \centering
    \includegraphics[width=\linewidth]{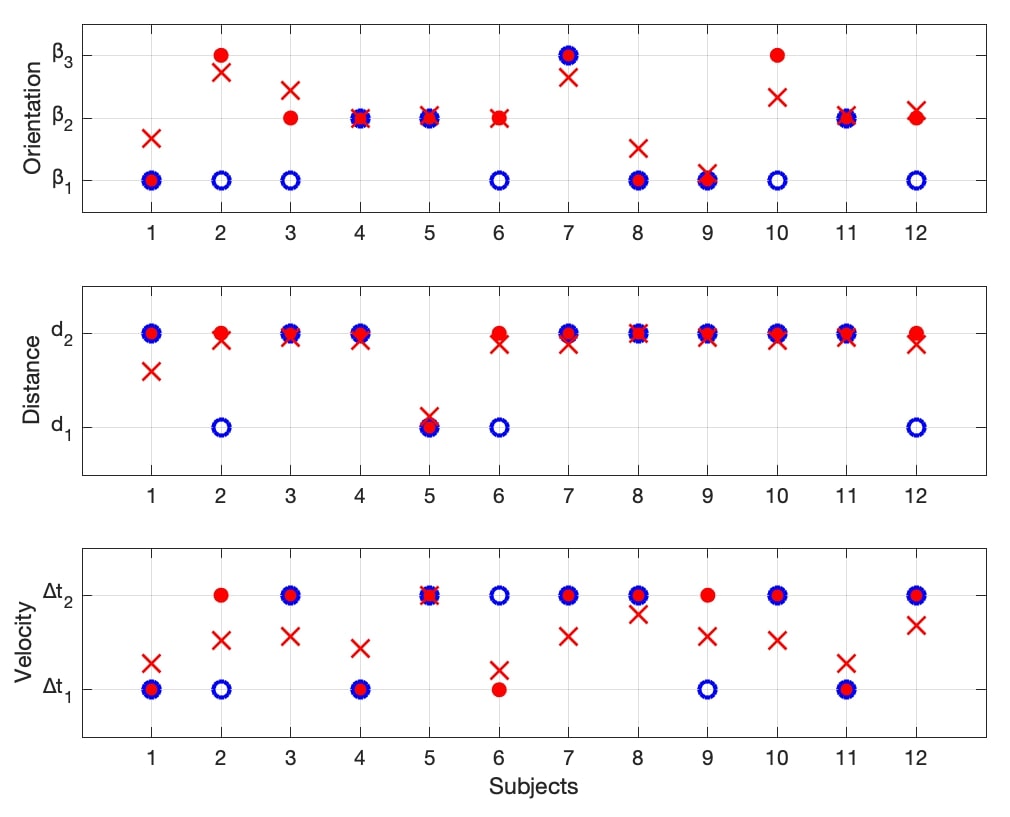}
    
    \vspace{-0.2cm}
    \caption{\small 
    Learned parameters (red full circles) and preferences (blue circles) for twelve subjects. Red crosses indicate the weighted average over last twenty-five steps. 
    }
    \label{fig:adaptedParams}
    \vspace{-0.7cm}
\end{figure}

Figure \ref{fig:adaptedParams} shows the result of the adaptation system, i.e. the learned mug's orientation, interaction distance and robot velocity to maximise human-robot \textit{coefficiency}, in comparison to the preferences stated by the twelve subjects involved in the experiments (\ref{RQ2}). 
In the plots, the full red circles represent the learned parameters, which are the most selected values (i.e. the variable mode) over the last twenty-five iterations of each experiment run. Since the algorithm keeps searching for the optimal value, we also computed the average parameter value in the same interval and reported it in the figure through red crosses. 
Finally, we depicted the preferred values indicated by each participant in the questionnaire as blue circles. 

Considering all subjects, at least two of the parameters reached convergence with the stated preferences within about $7$ minutes from the beginning of the interaction, which is, after $21.1$ iterations, on average. 
In this work, we consider the convergence achieved when 
the human-preferred value is selected by the policy most of the time (i.e. the mode over all iterations performed until then is the human-preferred value) and at least five times in a row. 
The learning converged to the preferred orientation for seven subjects and to the preferred interaction distance and robot velocity for nine out of twelve subjects (see Fig.\ref{fig:adaptedParams}). 
To determine the contribution of each parameter to a fruitful interaction, we computed the mean distance over all subjects between the learned parameter (red circle) and the average parameter value during the last iterations (red cross) and we obtained $0.13$, $0.10$, $0.33$ for orientation, distance, and velocity respectively.
The higher this distance, the lower the parameter contribution since it means the algorithm keeps jumping among its possible values. 
Besides, an average improvement of $4.6\%$ in the \textit{coefficiency} score was registered with preferred parameters ($p^{A_\text{pref}} = 0.07$). 

\vspace{-0.1cm}
\subsection{Subjective Questionnaires}
\vspace{-0.1cm}
Table \ref{tab:questionnaires} reports the results of post-study survey.
From the top half of the table, we can see that participants agreed that the robot motion's appropriateness and fluency should be taken into account. 
Interestingly, patterns of choices in the custom questionnaire indicate that participants consider the proposed costs equally relevant to plan well-coordinated behaviours. On average, the weights given to $C^\text{H}_\text{cognitive erg}$, $C^\text{H}_\text{physical erg}$, and $C^\text{R}_\text{energy cons}$ were $0.33$, $0.26$, and $0.41$, respectively. 

The bottom half of the table shows the Charalambous trust scores normalised in the interval $[1,5]$. Note that gripper-related items were removed. 
With respect to the reference values in \cite{Charalambous2016}, an increase of $110.7\%$ was registered for the perceived appropriateness (not normalised) of the robot motion and pick-up speed. 
Moreover, the proposed robot adaptation system slightly improved the perceived cooperation safety and reliability ($1.5\%$ and $3.7\%$, respectively). 
Overall, well-coordinated robot behaviours determined a predominant trust growth ($17.8\%$) in na$\ddot{\i}$ve participants.

\begin{table}[!t]
\caption{Results of post-study subjective questionnaires.}
\begin{tabularx}{\linewidth}
{p{6.5cm}p{0.5cm}p{0.5cm}}
    \toprule Custom Questionnaire 
     & Mean & Std \\ \midrule
    \textit{Q1}: The way the robot moved at the end of the experiment met my preferences.     & 3.58 & 1.00 \\
    \textit{Q2}: The robot behaved in an awkward and unnatural way.    & 2.25 & 1.22 \\
    \textit{Q3}: I felt comfortable while performing the task as I would be with another human.     & 4.17 & 0.83 \\ 
    \multicolumn{3}{p{8cm}}{\textit{Q4}: In planning how to configure the object and hand it over to you, the robot should take into account:} \\
    \quad \quad your mental load and perceived safety.  & 3.67 & 0.87  \\
    \quad \quad your physical effort.    & 3.33 & 1.41 \\
    \quad \quad the appropriateness and fluency of robot motion.      & 4.11 & 0.93 \\
    \bottomrule
\end{tabularx}

\smallskip

\begin{tabularx}{\linewidth}{p{6.5cm}p{0.5cm}p{0.5cm}}
    \toprule Charalambous Trust Scale 
     & Mean & Std \\ \midrule
    Perceived robot motion and pick-up speed     & 3.79 & 1.41 \\
    Perceived cooperation safety     & 4.31 & 0.73 \\
    Perceived robot reliability     & 4.25 & 1.06 \\
    \cmidrule(l{0.7em}r{1em}){1-3}
    Total     & 12.35 & 1.97 \\ \bottomrule
\end{tabularx}
\label{tab:questionnaires}
\vspace{-0.5cm}
\end{table} 

\vspace{-0.1cm}
\section{Discussion}
\label{sec:discussion}
\vspace{-0.1cm}

Results highlighted the ability of the proposed learning strategy to online maximise the benefits and minimise the effort of the specific agents involved in the cooperative task. 
Indeed, a statistically significant improvement in human cognitive and physical ergonomics was registered by exploiting the optimal interaction parameters learned by the system, and the robot expenses were noticeably reduced.
This means that we succeeded in embedding a concept of \textit{coefficiency} based on cognitive and physical factors inspired by theories of human joint actions into human-robot interactions.

Interestingly, acting \textit{coefficiently}, the robot was able to meet the individual preferences of most of the subjects who participated in the experiments. The proposed metrics of human comfort and discomfort (presented in Sec.\ref{sec:coefficiency}) were found to be appropriate for adjusting the robot interaction parameters on-the-fly and learning the personalised behaviour that best fits the user needs.  
However, not all parameters have the same impact on defining a fruitful interaction. 
Parameters more affecting human-robot \textit{coefficiency} are learned faster and more accurately, while less relevant ones may even not converge. 
As seen in Fig.\ref{fig:adaptedParams}, the mean distance over all subjects between the learned velocity and the average parameter value during the last iterations is higher than the ones obtained for the orientation and distance. We can deduce that the robot velocity is less relevant to the decision-making strategy.  

Nevertheless, measuring and expressing actual human preferences is not straightforward. 
For example, subject $5$ claimed that the robot did not facilitate his grasping even if the proposed object orientations evenly covered the operating space and parameters converged to the stated preferences. 
This makes us question the reliability of the self-reported values and complicates the evaluation of the system's accuracy. 

The main limitation of the framework is related to the assumptions in the definition of human-robot \textit{coefficiency}. For example, relying on the behavioural analysis in \cite{Kanda2003body}, we expect that participants divert attention from the mug when the interaction is annoying and not legible (i.e. the robot moving too slowly or performing an extensive rotation after the handover to come back to the homing configuration).
But, two participants exhibited behaviours far from our expectations thus preventing the system from appropriately learning. Subject $12$ forced herself to be overfocused and always performed the task in the same manner although the parameters were far from her preferences. Hence, the interaction parameters were learned only based on the robot expenses. On the contrary, subject $2$ tended to get distracted and protracted the motion initiation when the robot ran actions that were more legible and predictable for him. 
To solve these issues, we may extend the concept of \textit{coefficiency}, including more variables in addition to those currently exploited to fulfil learning inabilities encountered by the framework for some participants. 
Moreover, at this stage, we limited the search space's dimensionality given the potential countertrend of the efficiency costs, which may affect the convergence of the algorithm, and we exploited previous knowledge in the field of robot-to-human handovers. 
The promising results of this study encourage us to consider a wider range of interaction parameters in follow-up works and quantify the gain of capturing \textit{coefficiency} rather than any individual components optimisation through a within-subjects experiment. 
In general, the reader should note that this work does not aim to substantially improve the score \textit{per sé}. Still, it mainly focuses on presenting new metrics for a more natural interaction and providing initial proof of their potential utility.

Although the questionnaire revealed that, on average, subjects ranked the costs equally important to plan a seamless interaction, it would also be interesting to investigate the benefits of a subject-specific model of human-robot \textit{coefficiency} score. Assessing the weights that each subject would associate with each cost, we could define a reward function as a personalised weighted combination to address the individual demands and characteristics of the user. 

In addition, the subjective impressions reported in the questionnaires suggested that well-adapted robot behaviours are perceived as natural, improve the perceived motion appropriateness and foster trust in the automated partner with respect to the values provided in \cite{Charalambous2016} for early comparison. This remarkable outcome indicates that the proposed \textit{coefficiency} framework is a step forward on the way to developing robots that can be interacted with as easily as humans.

\section{Conclusions}
\label{sec:conclusion}

This study investigated whether transferring the human paradigm of acting \textit{coefficiently}, i.e. maximising the partner's benefits while being sensitive to own expenses, to human-robot cooperative tasks promotes a more seamless and natural interaction. 
We first modelled human-robot \textit{coefficiency} by detecting implicit indicators of human comfort and discomfort and computing the energy expended by the robot to accomplish the desired trajectory. 
Then, we proposed an RL approach to online adapt the robot behaviour, which exploits the human-robot \textit{coefficiency} score as a reward to learn the actions that maximise such \textit{coefficiency}. More specifically, the robot starts exploring by adjusting the value of different interaction parameters, then learns and selects the combination of them that ultimately best fits human preferences. 

The framework performed well for ten out of twelve participants, indeed at least an interaction parameter converged to the preferences specified in the questionnaires. However, occasional contradictory results raise doubts about the reliability of self-reported values and encourage us to investigate more variables related to human-body language and emotional cues presented in the literature. 
Future studies may also consider designing a subject-specific model of the reward function to cope with situations where our costs are not equally relevant or assumptions are not completely fulfilled. 
Moreover, it would be interesting to analyse the impact of variations of every parameter on the proposed costs and investigate the system adaptation over time. 

Overall, the adaptation mechanism developed in this study showed promising features to be applied in more complex collaborative tasks, involving, for instance, direct physical contact with the robot, and human whole-body movements and analysing stress-related motion patterns. 
Finally, the study of \textit{coefficiency} in human-robot handovers built the foundation for future applications of cognitive psychology to hybrid interaction settings.

\bibliographystyle{ieeetr}

\bibliography{bibliography_handovers}

\end{document}